\documentclass[sigconf]{acmart}

\AtBeginDocument{%
  }

\copyrightyear{2025}
\acmYear{2025}
\setcopyright{acmlicensed}\acmConference[WWW '25]{Proceedings of the ACM Web Conference 2025}{April 28-May 2, 2025}{Sydney, NSW, Australia}
\acmBooktitle{Proceedings of the ACM Web Conference 2025 (WWW '25), April 28-May 2, 2025, Sydney, NSW, Australia}
\acmDOI{10.1145/3696410.3714534}
\acmISBN{979-8-4007-1274-6/25/04}

\acmConference[Conference acronym 'XX]{Make sure to enter the correct
  conference title from your rights confirmation emai}{June 03--05,
  2018}{Woodstock, NY}
\acmISBN{978-1-4503-XXXX-X/18/06}

\usepackage{algorithm}
\usepackage{algorithmic}
\usepackage{appendix}
\usepackage{booktabs}
\usepackage{amsmath}
\usepackage{multicol}
\usepackage{multirow}
\usepackage{array}
\usepackage{bm}
\usepackage{pbox}
\usepackage{balance}
\usepackage{afterpage}
\usepackage{float} 
\usepackage{blindtext}
\usepackage{graphicx}

\graphicspath{{./graphics/}}
\DeclareGraphicsExtensions{.pdf,.png,.jpg}

\usepackage{multicol} 
\usepackage{makecell}
\newcolumntype{P}[1]{>{\centering\arraybackslash}p{#1}}

\usepackage{xcolor}

\newcommand\red[1]{\textcolor{red}{#1}}

\usepackage[belowskip=2pt,aboveskip=0pt]{caption}

\begin{document}

\title{Cross-Modal Transfer from Memes to Videos: Addressing Data Scarcity in Hateful Video Detection}

\author{Han Wang}
\orcid{0009-0007-4486-0693}
\affiliation{
  \institution{Singapore University of Technology and Design}
  \streetaddress{8 Somapah Road}
  \city{Singapore}
  \country{Singapore}
  \postcode{487372}
}
\email{han_wang@sutd.edu.sg}

\author{Rui Yang Tan}
\orcid{0009-0007-1325-5888}
\affiliation{
  \institution{Singapore University of Technology and Design}
  \streetaddress{8 Somapah Road}
  \city{Singapore}
  \country{Singapore}
  \postcode{487372}
}
\email{ruiyang_tan@sutd.edu.sg}

\author{Roy Ka-Wei Lee}
\orcid{0000-0002-1986-7750}
\affiliation{
  \institution{Singapore University of Technology and Design}
  \streetaddress{8 Somapah Road}
  \city{Singapore}
  \country{Singapore}
  \postcode{487372}
}
\email{roy_lee@sutd.edu.sg}

\renewcommand{\shortauthors}{Wang et al.}


\begin{abstract}

Detecting hate speech in online content is essential to ensuring safer digital spaces. While significant progress has been made in text and meme modalities, video-based hate speech detection remains under-explored, hindered by a lack of annotated datasets and the high cost of video annotation. This gap is particularly problematic given the growing reliance on large models, which demand substantial amounts of training data. To address this challenge, we leverage meme datasets as both a substitution and an augmentation strategy for training hateful video detection models. Our approach introduces a human-assisted reannotation pipeline to align meme dataset labels with video datasets, ensuring consistency with minimal labeling effort. Using two state-of-the-art vision-language models, we demonstrate that meme data can substitute for video data in resource-scarce scenarios and augment video datasets to achieve further performance gains. Our results consistently outperform state-of-the-art benchmarks, showcasing the potential of cross-modal transfer learning for advancing hateful video detection\footnote{Dataset and code available at \href{https://github.com/Social-AI-Studio/CrossModalTransferLearning}{https://github.com/Social-AI-Studio/CrossModalTransferLearning}.}. 

\red{Disclaimer: This paper contains sensitive content that may be disturbing to some readers.}

\end{abstract}




\maketitle
\section{Introduction}
\textbf{Motivation.} The rapid growth of social media platforms has intensified the need for effective hate speech detection methods. Such content, which targets individuals or groups based on attributes like race, ethnicity, nationality, religion, disability, age, veteran status, sexual orientation, or gender identity \cite{chhabra2023survey,schmidt2017survey}, poses significant risks to online communities. While extensive research has advanced hate speech detection in text \cite{davidson2017automated,razavi2010offensive,waseem2016hateful,zampieri2019predicting,xiao2024toxicloakcn,lee2024improving,awal2023model,cao2020hategan,awal2021angrybert} and meme modalities \cite{kiela2020hateful,pramanick2021detecting,maity2022multitask,fersini2022semeval}, video-based detection remains underexplored. Recent datasets like MultiHateClip (MHC) \cite{wang2024multihateclip} and HateMM \cite{das2023hatemm} have facilitated initial efforts in this space. However, these datasets are constrained by their scale, with approximately 1,000 annotated videos each—significantly fewer than text datasets such as HateSpeechDataset (24,802 samples) \cite{davidson2017automated} and meme datasets like Facebook Hateful Memes (10,000 samples) \cite{kiela2020hateful}.


The limited size of hate video datasets arises from two key challenges. First, stringent content moderation policies on platforms like YouTube significantly constrain the availability of hateful video content, drastically reducing the pool of videos suitable for analysis. For example, while MHC and HateMM initially gathered 5,000–6,000 videos, rigorous filtering to ensure label balance reduced their final datasets to approximately 1,000 videos each. Second, video annotation is inherently more labor-intensive than text or meme annotation. The multimodal nature and extended duration of videos require careful analysis of audio, visual, and textual cues, which makes the review process significantly slower, even when the annotation criteria are the same as those for text or memes.

\begin{figure}[t]
  \centering
  \includegraphics[width=0.45\textwidth]{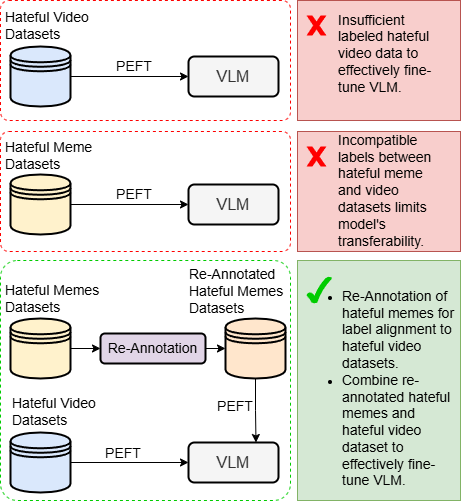}
  \caption{Comparison between our approach (bottom) and other solutions that performs Parameter-Efficient Fine-Tuning (PEFT) on VLM for hateful video detection.}
  \label{fig:pipeline}
\end{figure}

\textbf{Research Objectives.} Motivated by these challenges, we explore whether hateful meme datasets, which are more accessible and easier to annotate, can serve as effective substitute for training hateful video detection models. If successful, this approach could enable efficient dataset creation in scenarios with limited video but abundant meme data, such as bullying \cite{maity2022multitask} and misogyny detection \cite{fersini2022semeval}. Large language models (LLMs) and vision-language models (VLMs), which have shown impressive capabilities in detecting hate speech \cite{wang2023evaluating,hee2024recent}, hateful memes \cite{cao2023procap,cao2024modularized,hee2024bridging,zhu2022multimodal,cao2023prompting}, and videos \cite{wang2024multihateclip}, require vast amounts of training data to attain optimal performance. This study, therefore, investigates whether augmenting limited hateful video datasets with abundant hateful meme datasets can enhance model performance, addressing a critical gap in multimodal hateful video detection. Specifically, we conduct experiments using two meme datasets: Facebook Hateful Memes (FHM) \cite{kiela2020hateful} and Multimedia Automatic Misogyny Identification (MAMI) \cite{fersini2022semeval}, and two video datasets: MultiHateClip (MHC) \cite{wang2024multihateclip} and HateMM \cite{das2023hatemm}. Complementing these datasets, we utilized two state-of-the-art vision-language models: the image-based LLaMA-3.2-11B \cite{dubey2024llama} and the video-based LLaVA-NeXT-Video-7B \cite{zhang2024llava}.

Our initial experiments explored whether meme datasets could serve as effective substitute for training hateful video detection models. Fine-tuning models directly on meme datasets revealed performance limitations due to inconsistent label definitions across datasets. To address this, we developed a model-prediction-driven, human-assisted re-annotation pipeline to align meme dataset labels with video dataset definitions. This pipeline employed a majority voting framework, combining original meme labels, model predictions via few-shot prompting, and human annotations. The re-annotated meme datasets enabled models fine-tuned on meme data to achieve performance comparable to those fine-tuned on video data, demonstrating the viability of meme datasets as a substitute when video data is limited. Figure \ref{fig:pipeline} compares our approach and other potential hateful video detection solutions.

Building on these findings, we investigated whether re-annotated meme data could augment video datasets to further improve performance. Using the best-performing combination of the FHM dataset \cite{kiela2020hateful} and the LLaMA-3.2-11B model \cite{dubey2024llama}, we observed notable gains. Fine-tuning LLaMA-3.2-11B exclusively on video datasets improved Macro-F1 scores by up to 4\% for MHC \cite{wang2024multihateclip} and 3\% for HateMM \cite{das2023hatemm}, compared to without fine-tuning. Interestingly, fine-tuning on re-annotated FHM meme data achieved comparable or superior results to video fine-tuning alone. Combining FHM meme data with video datasets further enhanced performance, yielding an additional 3\% Macro-F1 improvement for MHC and 1\% for HateMM.

\textbf{Contributions.} We summarize our contributions as follows:
\begin{itemize}
    \item  We propose a novel, model-prediction-driven, human-assisted re-annotation method to align labels across meme and video datasets, ensuring greater label consistency. The re-annotated meme datasets are made publicly available to facilitate further research. 
    \item  We demonstrate that hateful meme datasets can effectively substitute for hateful video datasets in scenarios where annotated video data is unavailable, enabling the training of video detection models with comparable performance. 
    \item  We show that augmenting small video datasets with re-annotated hateful meme datasets significantly improves model performance, surpassing state-of-the-art results on the two benchmark video datasets, MHC and HateMM. 
\end{itemize}

\section{Related Work}
\textbf{Hate Speech Detection Across Modalities.} The detection of hate speech has been extensively studied in the text modality~\cite{cao2020deephate}, with numerous datasets curated from different social media platforms and forums~\cite{zhang2018detecting,ng2024sghatecheck,de-gibert-etal-2018-hate,gao2017detecting}. Most research broadly addresses hate speech classification, with some studies targeting specific forms like misogyny \cite{fersini2018overview, waseem-hovy-2016-hateful}, racism \cite{waseem-2016-racist}, or victim categorization \cite{warner2012detecting}, thus expanding the scope of text-based hate speech analysis.

Beyond text, the detection of hate speech has extended into multimodal domains, particularly memes. Notable datasets include FHM \cite{kiela2020hateful}, which focuses on multimodal hate meme detection, and others addressing harmful \cite{pramanick2021detecting}, cyberbullying \cite{maity2022multitask}, misogynistic content \cite{fersini2022semeval}, and war-related posts \cite{ng2024love}. These meme datasets, ranging from 3,000 to 10,000 samples, offer a robust foundation for developing multimodal models, especially for videos that share two key modalities with memes: text and vision. 

In contrast, hateful video detection remains under-explored, hindered by the scarcity of large, annotated datasets. Earlier works introduced smaller datasets, such as the 400 Portuguese YouTube videos compiled by \cite{alcantara2020offensive} and 300 English videos by \cite{wu2020detection}. Still, their limited size restricts the development of robust multimodal classification models. Recent advancements include HateMM \cite{das2023hatemm}, which compiles 1,083 BitChute videos for binary hate classification, and MHC \cite{wang2024multihateclip}, a multilingual dataset of 2,000 videos (1,000 English videos from YouTube and 1,000 Chinese videos from Bilibili). 

Despite these advancements, video-based hate speech datasets remain significantly smaller than text or meme datasets. This gap motivates our exploration of leveraging hateful meme datasets as effective substitutes or augmentations for video datasets in hate speech detection tasks. By utilizing the complementary nature of these modalities, our work aims to address the scarcity of video datasets while advancing multimodal classification research.

\textbf{Cross-Modal Transfer Learning.}
Cross-modality transfer learning has been widely studied, particularly in the transfer from images to videos. Many Vision Transformers, such as TimeSformer \cite{bertasius2021space_time_attention}, ViT \cite{dosovitskiy2021vit}, and Video-Swin \cite{liu2022video_swin}, are based on the Image Transformer architecture (ViT) pre-trained on large-scale image datasets like ImageNet. Pretraining on images enables these models to learn spatial features effectively. However, transferring knowledge from images to videos introduces the challenge of capturing temporal information. To address this, image-pretrained models typically undergo additional training on video datasets to model the temporal dynamics inherent in video content.

Most existing methods focus on fine-tuning within the same modality, with limited attention to cross-modality fine-tuning. Notable works, including \cite{carreira2017temporal}, \cite{pan2022st}, and \cite{yao2023side4video}, integrate spatial and temporal modeling to enhance image-to-video cross-modality transfer. While these methods achieve significant advancements, they primarily address the visual component of videos, often neglecting the multimodal nature of video content, which typically integrates visual, textual, and auditory modalities.

In contrast, transferring knowledge from memes to videos offers unique potential due to their shared multimodal nature. Memes inherently combine visual and textual elements, closely mirroring the multimodal structure of videos. By leveraging the social and cultural context embedded in meme content, this transfer approach could enhance video understanding across multiple modalities. Additionally, using hateful memes as substitute for hateful video datasets allows models to benefit from the rich, socially grounded information present in memes, providing a novel pathway for addressing the challenges of video-based hate speech detection.

\section{Methodology}

\definecolor{text}{rgb}{1.0, 0.251, 1.0}
\definecolor{question}{rgb}{0.310, 0.478, 0.157}
\definecolor{label}{rgb}{0.0, 0.337, 0.839}

\begin{figure}[t]
  \centering
  \includegraphics[width=0.48\textwidth]{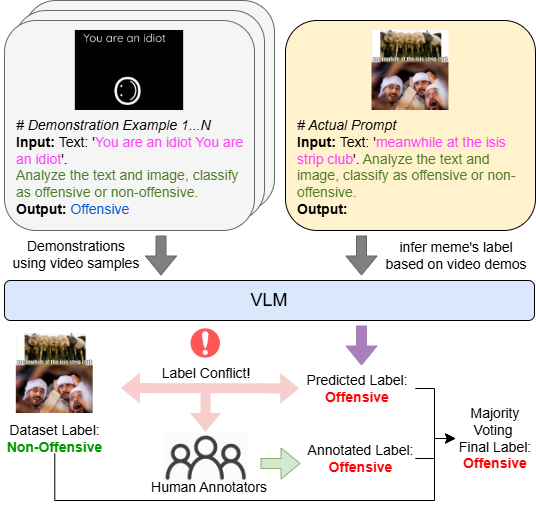}
  \caption{Re-annotation pipeline of meme datasets.}
  \label{fig:reannotation_pipeline}
\end{figure}


Our approach leverages hateful meme datasets as effective proxies for training hateful video detection models, addressing the challenges posed by the scarcity and high annotation cost of video datasets. This framework consists of two primary components: Meme Dataset Re-annotation and Fine-tuning. First, we address the challenge of label inconsistencies between meme and video datasets by introducing a re-annotation process. This process aligns the meme dataset's labels with the video dataset's label definitions using a majority voting mechanism, which combines the original dataset labels, model predictions, and human annotations. The re-annotated meme dataset ensures compatibility and consistency for downstream tasks. Next, we fine-tune a VLM on the re-annotated meme dataset for hateful meme classification. The fine-tuned model is evaluated on hateful video classification tasks, demonstrating the effectiveness of using memes as a substitute for videos in multimodal classification. We also explore augmenting the original videos with re-annotated memes to fine-tune VLMs to improve hateful video detection performance.

To provide a comprehensive understanding of our methodology, Subsection~\ref{subsec:formulation} defines the problem formulation, Subsection~\ref{subsec:reannotation} details the re-annotation process, and Subsection~\ref{subsec:finetuning} explains the cross-modality fine-tuning methodology.




\subsection{Problem Formulation}
\label{subsec:formulation}

We define the hateful video detection task as classifying a given video \(V\) into two categories: hateful/offensive (\(y=1\)) or non-hateful/non-offensive (\(y=0\)). The HateMM dataset provides binary labels: hateful and non-hateful. In contrast, the MHC dataset includes hateful, offensive, and normal labels. To harmonize label definitions across datasets, we merge the hateful and offensive categories in MHC into a single offensive label, resulting in a binary classification setup: offensive (\(y=1\)) vs. non-offensive (\(y=0\)).

Each video comprises three modalities: text (\(T\)), audio (\(A\)), and vision (\(V\)). The text modality is represented as a sequence of words \(\{w_1, w_2, \ldots, w_m\}\), aggregated from the video’s title (if available) and transcript. The audio modality captures the auditory content of the video, while the vision modality is represented by a sequence of frames \(\{f_1, f_2, \ldots, f_k\}\) extracted from the video.

Similarly, each meme includes two modalities: text (\(T\)) and vision (\(V\)). The text modality is a sequence of words \(\{w_1, w_2, \ldots, w_m\}\) derived from the meme’s textual content, and the vision modality consists of a single image representing the meme’s visual content.

Given the limited availability of large-scale Vision-Audio-Language Models, this study focuses on the \textbf{Text} (\(T\)) and \textbf{Vision} (\(V\)) modalities, using VLMs as the base model \(f\). The goal is to fine-tune \(f\) to map these two modalities to a ground-truth binary label \(y\), such that:

\begin{equation}
\label{eq:formulation}
f: (T, V) \rightarrow y, \quad y \in \{0, 1\}.
\end{equation}

By unifying label definitions and leveraging multimodal inputs, this formulation ensures consistency across datasets and enables the application of VLMs for effective hateful video classification.




\subsection{Re-Annotation for Label Alignment}
\label{subsec:reannotation}


Effectively aligning labels between meme and video datasets is a critical step in our approach. The re-annotation process integrates three key sources: the original dataset label, model predictions, and human annotations, as illustrated in Figure~\ref{fig:reannotation_pipeline}. This majority voting-based approach ensures label consistency and improves compatibility across datasets.

\subsubsection{\textbf{Overview of the Re-Annotation Process}} The re-annotation pipeline is designed to address discrepancies between meme dataset labels and video dataset definitions. The original meme labels are remapped to align with the video dataset’s labelling scheme. For example, the FHM dataset has hateful and non-hateful labels, while the MHC dataset uses offensive and non-offensive labels. To align them, FHM's hateful is mapped to offensive, and non-hateful to non-offensive. No changes are needed to align FHM with HateMM, as both use the same hateful vs. non-hateful labels. For the MAMI dataset, misogynous maps to offensive for MHC and hateful for HateMM, while non-misogynous maps to non-offensive for MHC and non-hateful for HateMM. This remapped version is referred to as the original meme dataset in subsequent experiments.

To enhance label alignment, we leverage VLMs to generate predictions using few-shot prompting. These predictions leverage examples sampled from the video datasets, guiding the model to classify memes according to video label definitions. When the dataset label and model prediction conflict, human annotators with expertise in hate speech detection—resolve the disagreement by annotating based on the video dataset’s definitions. The final label for each meme is determined by majority voting among the three sources: dataset label, model prediction, and human annotation.

\begin{table}[t]
\centering
\small
\caption{Few-shot evaluation results for MHC and HateMM datasets using the non-finetuned LLaMA-3.2-11B and LLaVA-NeXT-Video-7B. The best-performing results for each dataset and model are bolded. Metrics: Accuracy (Acc) and Macro-F1 (M-F1).}
\begin{tabular}{c|c|c|c|c|c}
\hline
\multirow{2}{*}{\textbf{Model}} & \multirow{2}{*}{\textbf{\textit{N}}} & \multicolumn{2}{|c}{\textbf{MHC}} & \multicolumn{2}{|c}{\textbf{HateMM}}  \\ \cline{3-6} 
 & & \textbf{Acc} & \textbf{M-F1} & \textbf{Acc} & \textbf{M-F1}   \\ \hline \hline
\multirow{5}{*}{LLaMA-3.2-11B} & 0 & 0.66& 0.62& \textbf{0.79} & \textbf{0.78}\\ \cline{2-6} 
&2& \textbf{0.79}& \textbf{0.74}& 0.76 & 0.76\\ \cline{2-6} 
&4& 0.77& 0.74& 0.79 & 0.78\\ \cline{2-6} 
&6& 0.79 & 0.74& 0.79 & 0.78\\ \cline{2-6} 
&8& 0.77& 0.73& 0.78 & 0.78\\ \hline

\multirow{5}{*}{LLaVA-NeXT-Video-7B} & 0& 0.54& 0.53& 0.67 & 0.66\\ \cline{2-6} 
&2& \textbf{0.69}& \textbf{0.69}& 0.70 & 0.70\\ \cline{2-6} 
&4& 0.64& 0.63& \textbf{0.73} & \textbf{0.73}\\ \cline{2-6} 
&6& 0.66& 0.65& 0.70 & 0.69\\ \cline{2-6} 
&8& 0.62& 0.62& 0.65 & 0.64\\ \hline

\end{tabular}
\label{tab:llama_few_shots}
\end{table}

\subsubsection{\textbf{Model Prediction with Few-Shot Prompting}}
We employed two state-of-the-art VLMs: LLaMA-3.2-11B \cite{dubey2024llama} and LLaVA-NeXT-Video-7B \cite{zhang2024llava}, leveraging few-shot prompting to guide the models for re-annotation. In the case of LLaMA-3.2-11B, the prompting process includes multiple real video demonstrations, each containing vision features, \textcolor{text}{text}, a \textcolor{question}{question}, and the corresponding \textcolor{label}{label}, as shown in Figure~\ref{fig:reannotation_pipeline}. For the queried memes, prompts followed the same structure but excluded the label, which the model was tasked to predict. LLaVA-NeXT-Video-7B, on the other hand, does not support multiple demonstrations. Instead, vision-based descriptions of video content provided by human annotators replaced the visual input in the demonstration examples.

To determine the optimal number of demonstration examples (\textbf{\textit{N}}) for model prediction, we evaluated prompting with 0, 2, 4, 6, and 8 video examples, using the MHC \cite{wang2024multihateclip} and HateMM \cite{das2023hatemm} datasets for demonstration and testing. This evaluation was conducted on both LLaMA-3.2-11B and LLaVA-Next-Video-7B, as shown in Table~\ref{tab:llama_few_shots}.

The results revealed diminishing returns beyond a certain threshold for \textbf{\textit{N}}. For example, on MHC with LLaMA-3.2-11B, the Macro-F1 score improved by 0.12 when \textbf{\textit{N}} increased from 0 to 2 but showed negligible gains from 2 to 8. This finding suggests that demonstration examples primarily aid task comprehension rather than significantly enhancing the model’s ability to discriminate between (\(y=1\))  and (\(y=0\)) content. Interestingly, LLaMA-3.2-11B achieved its best performance on the HateMM dataset in a zero-shot setting, with an M-F1 of 0.78, closely approaching the benchmark M-F1 of 0.79 reported in the original dataset paper. The strong zero-shot performance could be due to potential data leakage, as LLaMA-3.2-11B was released after the HateMM dataset. Although the authors of the model did not explicitly state that it was trained on HateMM, this possibility cannot be ruled out.

Between the two models, LLaMA-3.2-11B demonstrated superior performance due to its larger model size and ability to incorporate multiple demonstrations effectively. For the re-annotation of the meme dataset, we utilized the optimal \textbf{\textit{N}} value of video demonstrations and selected LLaMA-3.2-11B as the prediction model to generate labels. This approach ensured that predictions closely aligned with the video dataset’s label definition, forming a robust foundation for the subsequent majority voting process.

\begin{figure}[t]
  \centering
  \includegraphics[width=0.48\textwidth]{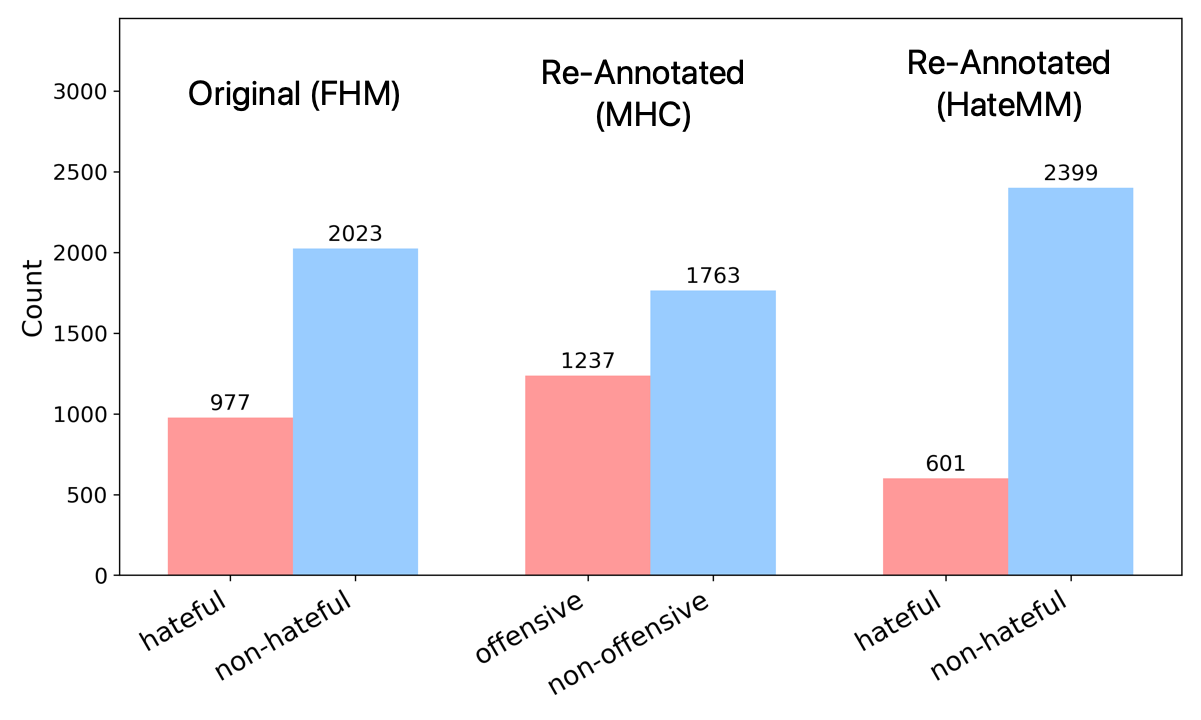}
  \caption{Meme count distribution for FHM dataset, categorized by original labels, re-annotated labels by MHC, and HateMM definitions.}
  \label{fig:fhm_statistics}
\end{figure}

\begin{figure}[t]
  \centering
  \includegraphics[width=0.48\textwidth]{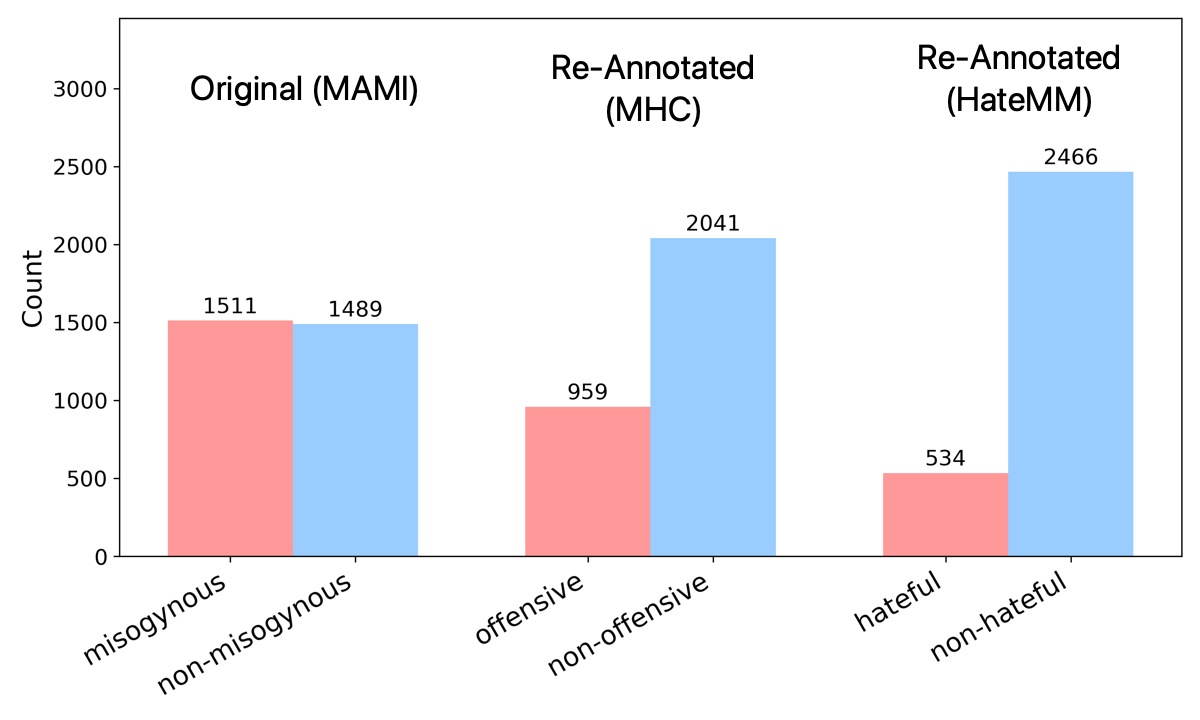}
  \caption{Meme count distribution for MAMI dataset, categorized by original labels, re-annotated labels by MHC, and HateMM definitions.}
  \label{fig:MAMI_statistics}
\end{figure}

\subsubsection{\textbf{Analysis of Re-Annotation Results}}
The re-annotation process yielded significant insights into label alignment and the challenges inherent in hateful speech detection. Figures~\ref{fig:fhm_statistics} and~\ref{fig:MAMI_statistics} illustrate the data distribution for FHM and MAMI original meme dataset labels alongside the re-annotated labels aligned with the video dataset definitions. These results reveal notable discrepancies, highlighting the necessity of re-annotation for consistent labelling.

Two primary factors contributed to these discrepancies. First, differences in label definitions between datasets played a major role. For instance, the label definitions in the MAMI dataset differ more significantly from those in video datasets compared to the FHM dataset, resulting in greater label discrepancies. Additionally, the inherently subjective nature of hateful speech detection exacerbates annotation disagreements, even among human annotators. For example, despite FHM and HateMM sharing the same definitions, label differences persist. The discrepancies between original and re-annotated labels highlight the importance of integrating human expertise to resolve ambiguities. By ensuring alignment with video dataset definitions, the re-annotation process enhances the comparability of the meme datasets, forming a robust foundation for subsequent experiments.

\subsection{Cross-Modality Fine-Tuning}
\label{subsec:finetuning}

Fine-tuning memes and testing videos introduce the challenge of aligning modalities between these two data types. The text modality is processed similarly for both memes and videos. The input to the VLM includes the text content of the meme or video, a corresponding question, and the ground truth label. However, the vision modality requires different handling, as videos consist of multiple frames, while memes are represented by single images.

To address this disparity, the alignment strategy is tailored to the supported modalities of the VLM. For image-based VLMs, such as LLaMA-3.2-11B \cite{dubey2024llama}, which accept only single images, the vision modality of a video is represented by randomly sampling one frame. For video-based VLMs, such as LLaVA-NeXT-Video-7B \cite{zhang2024llava}, which can process multiple frames, we sample 16 frames per video, aligning with the model author's recommendations and the average length of our video dataset. To adapt meme data for video-based VLMs, we simulate a video format by applying random image augmentations (e.g., rotation, cropping, etc.) to generate 16 frames from a single meme image. We employ Low-Rank Adaptation (LoRA) \cite{hu2021lora} adapters to fine-tune the pre-trained VLM during. Specifically, we insert LoRA adapters into all the query and value layers within the attention mechanism while keeping the initial layers frozen. This approach ensures efficient adaptation with minimal computational overhead, as only the LoRA layers are fine-tuned.

We optimize the model using the Cross Entropy Loss (\(\mathcal{L}_{\text{cross}}\)), designed for sequence-to-sequence learning tasks. Let \( m' \) denote the number of words in the input text (meme/video text and question), and \( k \) represent the number of frames (1 for LLaMA-3.2-11B and 16 for LLaVA-NeXT-Video-7B). The input consists of text tokens \(\{w_1, w_2, \ldots, w_{m'}\}\) and vision features \(\{f_1, f_2, \ldots, f_k\}\), with the ground truth label represented as \( w_{GT} \). The target output is the sequence \(\{w_1, w_2, \ldots, w_{m'}, w_{GT}\}\). The loss function is defined as:

\begin{equation}
\label{eq:loss}
\mathcal{L}_{\text{cross}} = - \sum_{t=1}^{m'+1} \log P(w_t \mid \{w_1, w_2, \ldots, w_{t-1}\}, \{f_1, f_2, \ldots, f_k\})
\end{equation}

Here, \( P(w_t \mid \{w_1, w_2, \ldots, w_{t-1}\}, \{f_1, f_2, \ldots, f_k\}) \) denotes the probability of predicting the word \( w_t \) in the target sequence, conditioned on the preceding words and frames. This formulation aligns the text and vision modalities to optimize cross-modality performance, ensuring effective fine-tuning for meme and video datasets.

\section{Experiments}
\label{sec:experiments}
This section outlines our experimental framework for evaluating the cross-modal transfer from hateful memes to hateful video detection. We describe the datasets, Vision Language Models (VLMs), and fine-tuning strategies employed, as detailed in Section~\ref{setting}. Next, we perform a hyperparameter analysis to determine the optimal amount of training data required for effective cross-modal transfer. Following this, we present the results for two key scenarios: using memes as substitutes for video datasets and using memes as augmentations to enhance video dataset performance. Finally, we conduct an error analysis to provide illustrative examples of real-world predictions under different fine-tuning strategies.

The experiments address two core research questions: 

\begin{itemize} 
    \item \textbf{RQ1}: Can hateful meme datasets effectively substitute hateful video datasets in training VLMs when video data is scarce or unavailable? 
    \item \textbf{RQ2}: Can hateful meme datasets augment video datasets to improve the performance of VLMs beyond what is achievable with video data alone? 
\end{itemize}

The findings related to \textbf{RQ1} and \textbf{RQ2} are presented in Sections~\ref{substitution} and~\ref{augmentation}, respectively. Section~\ref{error} concludes the experimental analysis by showcasing examples of predictions on real-world video data, offering deeper insights into the efficacy of different fine-tuning strategies.



\subsection{Experimental Settings}
\label{setting}
\subsubsection{\textbf{Datasets}}
Our experiments utilize two video datasets — MHC\cite{wang2024multihateclip} and HateMM \cite{das2023hatemm} — and two meme datasets — FHM \cite{kiela2020hateful} and MAMI \cite{fersini2022semeval}. Key details of these datasets are as follows:

\begin{itemize}
    \item \textbf{MHC}: This multilingual dataset includes 1,000 English YouTube videos categorized into hateful (82), offensive (256), and normal (662). For binary classification, we merge the hateful and offensive categories, yielding 338 offensive and 662 non-offensive videos. The offensive label represents ``\textit{content that could cause discomfort or distress.}''
    \item \textbf{HateMM}: Comprising 1,083 videos from BitChute (431 hateful, 652 non-hateful), this dataset defines hateful content as ``\textit{targeting groups or individuals based on characteristics such as religion, ethnicity, origin, sexual orientation, gender, physical appearance, disability, or disease.}''
    \item \textbf{FHM}: This meme dataset includes 9,000 samples (3,300 hateful, 5,700 non-hateful) from Facebook, with the hateful label reflecting the same hate speech definition as HateMM.
    \item \textbf{MAMI}: Consisting of 8,795 memes (3,300 misogynous, 5,495 non-misogynous), this dataset also originates from Facebook, with the misogynous label representing ``\textit{content targeting women with hateful messages.}''
\end{itemize}

\subsubsection{\textbf{Models}}
We employ two state-of-the-art VLMs in this study: LLaMA-3.2-11B \cite{dubey2024llama} and LLaVA-NeXT-Video-7B \cite{zhang2024llava}. Both models have demonstrated exceptional performance in tasks such as visual question answering (VQA) and image/video captioning. LLaMA-3.2-11B is an image-based model designed to process single images but lacks support for video input. In contrast, LLaVA-NeXT-Video-7B is a video-based model capable of handling video input by processing multiple image frames, making it particularly suited for multimodal tasks involving temporal dynamics. Our experiments aim to evaluate the consistency of meme-based performance across these two distinct model types, highlighting their strengths and limitations in cross-modal transfer for hateful video detection.


\subsubsection{\textbf{Fine-Tuning Strategies}}
Our experiments employed five fine-tuning strategies to evaluate cross-modal transfer:  
\begin{itemize}
    \item No Fine-Tune \textbf{(No FT)}: The VLM performs inference on the video dataset without fine-tuning.  
    \item Video Fine-Tune \textbf{(Vid-FT)}: The VLM is fine-tuned on the video dataset's training set before testing on its test set.  
    \item Original Meme Fine-Tune \textbf{(OM-FT)}: The VLM is fine-tuned on the original meme dataset and tested on the video dataset.  
    \item Re-annotated Meme Fine-Tune \textbf{(RM-FT)}: The VLM is fine-tuned on the re-annotated meme dataset and tested on the video dataset.  
    \item Video + Re-annotated Meme Fine-Tune \textbf{(Vid+RM-FT)}: The VLM is fine-tuned on a video dataset augmented with the re-annotated meme dataset before testing on the video dataset.  
\end{itemize}

For video datasets, we adopted an 80/20 training-to-testing split, as specified in the original dataset papers. For meme datasets, we randomly sampled 3,000 memes for fine-tuning. Fine-tuning was performed over 5 epochs with a batch size of 8 and a learning rate of 2e-4. Testing on videos for \textbf{(No FT)} employed few-shot prompting with the optimal number of demonstrations (\textbf{\textit{N}}) identified earlier, while other strategies were evaluated using zero-shot prompting, consistent with the fine-tuning approach. The best-performing epoch on the video test data set was selected to report the results.



\subsection{Hyperparameter Analysis}
\begin{table}[t]
\centering
\small
\caption{Evaluation results of fine-tuning LLaMA-3.2-11B with different dataset sizes (\textbf{\textit{n}}) from the re-annotated FHM and MHC datasets, tested on the MHC dataset. The best-performing results for each dataset are bolded. Metrics: Accuracy (Acc) and Macro-F1 (M-F1).}
\begin{tabular}{c|c|c|c|c}
\hline
\multicolumn{1}{p{1.5cm}|}{\textbf{Fine-tune Dataset}}  & \textbf{\textit{n}} & \multicolumn{1}{p{1.5cm}|}{\textbf{Annotation Time (h)}} & \textbf{Acc} & \textbf{M-F1}    \\ \hline 
\hline
MHC &  200 & 13.3 & 0.68 & 0.67 \\ \hline
MHC & 400 & 26.7 & 0.69 & 0.68 \\ \hline
MHC & 600 & 40.0  & 0.77 & 0.76 \\ \hline
MHC & 800 & 53.3 & \textbf{0.79} & \textbf{0.78} \\ \hline \hline

FHM & 1000& 3.5  & 0.74 & 0.74 \\ \hline
FHM &2000 & 7.0 & 0.80 & 0.78 \\ \hline
FHM &3000& 10.5 & \textbf{0.81} & \textbf{0.80} \\ \hline
FHM &4000 & 14.0 & 0.80 & 0.79 \\ \hline
\end{tabular}
\label{tab:ablation_study}
\end{table}

To determine the optimal number (\textbf{\textit{n}}) of memes and videos for fine-tuning, we sought to balance annotation cost (in terms of time) with model performance. Fine-tuning experiments were conducted using LLaMA-3.2-11B on the re-annotated FHM and MHC datasets, with testing performed on the MHC dataset. Table~\ref{tab:ablation_study} summarizes the evaluation results and the corresponding annotation times. We tested meme dataset sizes of 1,000, 2,000, 3,000, and 4,000 samples, and video dataset sizes of 200, 400, 600, and 800 samples.

\textbf{Performance vs. Dataset Size.} Our experiments revealed diminishing returns in performance as dataset sizes exceeded specific thresholds. The threshold was reached with 600 samples for video data and 2,000 samples for meme data. To strike a balance between efficiency and performance, we selected 800 videos (the maximum available) and 3,000 memes (for optimal performance) for subsequent experiments. These dataset sizes for meme data delivered the best overall results on the MHC hateful video detection task, validating the effectiveness of using memes as a cost-efficient augmentation strategy for video datasets.

\textbf{Annotation Time.}
Annotation time is a critical consideration in this study, as it directly influences the feasibility of expanding training datasets for hateful video detection. Annotating a short video (approximately 1 minute) is estimated to take about 2 minutes, as per our experiments and the HateMM dataset paper \cite{das2023hatemm}. This process requires at least two annotations per video to ensure accuracy, making video annotation labor-intensive and time-consuming. In contrast, annotating a single meme in the FHM dataset takes approximately 0.5 minutes by an expert researcher. Moreover, for meme re-annotation, human evaluation is required only when there is disagreement between the dataset label and the model prediction. In the FHM dataset, only 42\% of memes needed re-annotation under the MHC label definitions. As a result, meme re-annotation is significantly more time-efficient, costing roughly one-quarter of the time required for video annotation.

This stark difference in annotation cost highlights the value of leveraging memes to augment hateful video detection. By utilizing the relatively cheaper and faster annotation process for memes, we can effectively expand training datasets while minimizing resource constraints. This advantage is even more pronounced for datasets like HateMM, where videos average 2–3 minutes in duration, further emphasizing the scalability of our approach.

\subsection{Meme as Video Substitution}
\label{substitution}

\begin{table*}[htbp]
  \centering
  \small
     \caption{Evaluation results of four fine-tuning strategies and testing on the MHC and HateMM datasets using LLaMA-3.2-11B and LLaVA-Next-Video-7B. The best-performing results for each dataset and each model are bolded. Metrics: Accuracy (Acc), Macro-F1 (M-F1), F1, Recall (R), and Precision (P). Classes: O (offensive), H (hateful).}
  \begin{tabular}{lccccccc|ccccc}
    \toprule 
    &&&  \multicolumn{5}{c}{\textbf{MHC}}  & \multicolumn{5}{c}{\textbf{HateMM}}\\
    \textbf{Model}&\textbf{Strategy}& \textbf{Fine-tune Dataset}  & \textbf{Acc} & \textbf{M-F1} & \textbf{F1(O)} & 
     \textbf{R(O)}  & \textbf{P(O)} & \textbf{Acc}  &\textbf{M-F1} & \textbf{F1(H)} & \textbf{R(H)}  & \textbf{P(H)}  \\
    \midrule\hline
LLaMA-3.2-11B &No FT&-  & 0.79 & 0.74 & 0.63 & 0.55 & \textbf{0.74}  & 0.79& 0.78& 0.75& 0.72& \textbf{0.78} \\  
LLaMA-3.2-11B &Vid-FT &MHC  & 0.79 & 0.78& 0.73& 0.87& 0.63& -& -& -& -& - \\  
LLaMA-3.2-11B &Vid-FT &HateMM  & - & -& -& -& -& \textbf{0.81}& \textbf{0.81}& \textbf{0.79}& 0.82& 0.77 \\  
LLaMA-3.2-11B &OM-FT  &FHM & 0.77 & 0.73 & 0.62 & 0.58& 0.67 & 0.80& 0.80& \textbf{0.79}& \textbf{0.85}& 0.74 \\  
LLaMA-3.2-11B &RM-FT &FHM  & \textbf{0.81} & \textbf{0.80}& \textbf{0.75}& \textbf{0.88}& 0.66& \textbf{0.81} & \textbf{0.81}& \textbf{0.79}& 0.83& 0.75 \\  
LLaMA-3.2-11B &OM-FT &MAMI  & 0.67 & 0.63  & 0.50 & 0.51 & 0.50 & 0.76  & 0.75 & 0.70& 0.64& \textbf{0.78} \\  
LLaMA-3.2-11B &RM-FT &MAMI  & 0.80 & 0.79& 0.74& 0.85& 0.66& 0.79& 0.79& 0.78& 0.83& 0.73 \\  
\hline
LLaVA-Next-Video-7B & No FT & - & 0.69 & 0.69& 0.67& \textbf{0.94} & 0.52& 0.73& 0.73& 0.74& 0.86& 0.64 \\  
LLaVA-Next-Video-7B & Vid-FT&MHC & 0.78 & 0.76 & 0.71& 0.81 & 0.63& -& -& -& -& - \\  
LLaVA-Next-Video-7B & Vid-FT&HateMM  & - & -& -& - & - & \textbf{0.80} &  \textbf{0.80} & \textbf{0.77} & 0.76 & 0.78 \\  
LLaVA-Next-Video-7B & OM-FT &FHM & 0.72 & 0.70 & 0.63 & 0.72 & 0.56 & \textbf{0.80} &  0.79 & 0.76 & 0.73 & \textbf{0.80} \\  
LLaVA-Next-Video-7B & RM-FT &FHM & \textbf{0.81} & \textbf{0.79} & 0.71 & 0.67 & \textbf{0.75} & \textbf{0.80} &  \textbf{0.80} & \textbf{0.77} & 0.77 & 0.78 \\  
LLaVA-Next-Video-7B & OM-FT  &MAMI  & 0.57 & 0.57  & 0.59 & 0.91 & 0.44 & 0.61  & 0.59 & 0.68 & \textbf{0.96} & 0.53\\  
LLaVA-Next-Video-7B & RM-FT &MAMI & 0.80 & 0.78 & \textbf{0.72} & 0.78 & 0.67 &  0.73 & 0.73 & 0.74 & 0.90 & 0.63 \\  

    \bottomrule
  \end{tabular}
 \label{tab:llama_results}
\end{table*}


To evaluate the potential of memes as effective substitutes for video datasets in training hateful video detection models, we fine-tuned the VLM using four strategies: \textbf{No FT}, \textbf{Vid-FT}, \textbf{OM-FT}, and \textbf{RM-FT}. Our experiments aimed to address \textbf{RQ1}, which explores whether memes, particularly after re-annotation to align with video dataset definitions, can serve as proxies for video datasets and achieve good classification performance. The models were evaluated on multiple metrics, including Accuracy (Acc), Macro-F1 (M-F1), F1, Recall, and Precision for the \(y=1\) label, with results presented in Table~\ref{tab:llama_results}.

The findings highlight several important insights. First, despite processing only single images, LLaMA-3.2-11B outperformed LLaVA-NeXT-Video-7B in Acc and M-F1 across dataset and fine-tuning strategies, highlighting the significance of model size and capacity in cross-modal transfer tasks. Second, fine-tuning on the FHM dataset consistently outperformed the MAMI dataset. This can be attributed to FHM’s broader focus on hateful speech, which aligns more closely with the hate video detection task, compared to MAMI’s specific focus on misogyny. These findings underscore the importance of selecting meme datasets that align closely with the target task when leveraging memes for cross-modal transfer.

Crucially, the re-annotation of meme datasets proved to be a key factor in improving performance. Fine-tuning on the re-annotated dataset (\textbf{RM-FT}) significantly enhanced performance compared to the original dataset (\textbf{OM-FT}), approaching or even exceeding the performance of models fine-tuned directly on video datasets (\textbf{Vid-FT}). This demonstrates the value of aligning meme dataset labels with video task definitions to improve transferability. Task alignment is crucial, as demonstrated by the poor performance when fine-tuning on the re-annotated MAMI dataset and testing on HateMM, where the task definitions differ. Notably, fine-tuning on re-annotated MAMI resulted in strong performance on MHC, likely because MHC focuses on gender-based hate speech, which is more similar to MAMI’s task of detecting misogyny. These results suggest that memes, particularly when re-annotated, can effectively substitute for video datasets in tasks with shared label definitions or domain-specific overlaps.

Overall, our findings emphasize the potential of memes as scalable and cost-effective proxies for video datasets in multimodal hate speech detection. By leveraging re-annotated meme datasets that align closely with the target task, models can achieve performance comparable to, or even surpassing, those trained directly on video datasets. This approach addresses the challenges of limited video data availability and high annotation costs, offering a practical solution for resource-constrained multimodal classification tasks.

\subsection{Meme as Video Augmentation}
\label{augmentation}

\begin{table*}[t]
\centering
\small
\caption{Evaluation results of fine-tuning with augmented video datasets using the re-annotated FHM dataset, tested on video datasets, compared to fine-tuning on video and meme data separately, and the best results from the video dataset paper. The best-performing results for each dataset are bolded. Metrics: Accuracy (Acc), Macro-F1 (M-F1), F1, Recall (R), and Precision (P).
    Classes: O (offensive), H (hateful).}
\begin{tabular}{lccccc|ccccc}
    \toprule
    & \multicolumn{5}{c}{\textbf{MHC}}  & \multicolumn{5}{c}{\textbf{HateMM}}\\
    \textbf{Strategy}&   \textbf{Acc} & \textbf{M-F1} & \textbf{F1(O)} & 
     \textbf{R(O)}  & \textbf{P(O)} & \textbf{Acc}  &\textbf{M-F1} & \textbf{F1(H)} & \textbf{R(H)}  & \textbf{P(H)}  \\
    \midrule \hline
Best Results from MHC paper \cite{wang2024multihateclip} & 0.81 & 0.79 & 0.73 & 0.72 &  \textbf{0.73} & - & - & - & - & -\\ 
Best Results from HateMM paper \cite{das2023hatemm}  & - & - & - & - & - & 0.80 & 0.79 & 0.75& 0.74 & 0.76 \\
Vid-FT   & 0.79 & 0.78& 0.73& 0.87 & 0.63& 0.81& 0.81& 0.79& 0.82 & 0.77 \\  
RM-FT & 0.81 & 0.80 & 0.75 & \textbf{0.88}& 0.66& 0.81 & 0.81 & 0.79 & \textbf{0.83}& 0.75 \\  

Vid+RM-FT & \textbf{0.82} & \textbf{0.81} & \textbf{0.76}  & 0.87 & 0.68 & \textbf{0.82} & \textbf{0.82} & \textbf{0.80} & 0.80 &  \textbf{0.79} \\ 
\hline
\end{tabular}
\label{tab:final_results}
\end{table*}
 
To assess the effectiveness of memes as augmentations for video datasets (\textbf{RQ2}), we employed the \textbf{Vid+RM-FT} strategy, which combines video datasets with the re-annotated meme dataset for fine-tuning. Based on the strong performance of \textbf{RM-FT} using the FHM dataset and LLaMA-3.2-11B across both video datasets (Table~\ref{tab:llama_results}), we chose FHM as the re-annotated meme dataset and LLaMA-3.2-11B as the base model for these experiments. Augmented fine-tuning was conducted on the HateMM and MHC datasets, with results compared against separate fine-tuning on video (\textbf{Vid-FT}), meme (\textbf{RM-FT}), and the best results from the video dataset paper. The results, presented in Table~\ref{tab:final_results}, demonstrate that \textbf{Vid+RM-FT} consistently outperformed the other strategies.

Data augmentation with \textbf{Vid+RM-FT} improved M-F1 by 0.03 for MHC and 0.01 for HateMM compared to \textbf{Vid-FT}. A comparative analysis revealed notable differences in predictions between \textbf{Vid+RM-FT} and \textbf{Vid-FT}. For MHC, \textbf{Vid+RM-FT} corrected 11 misclassified instances while introducing 4 errors that \textbf{Vid-FT} had classified correctly. Similarly, for HateMM, \textbf{Vid+RM-FT} corrected 7 instances and introduced 5 errors compared to \textbf{Vid-FT}. These findings indicate that augmenting video datasets with re-annotated memes significantly influences prediction and improves overall test performance, reducing errors in certain contexts.

While \textbf{Vid+RM-FT} outperform the baseline methods for both datasets, the improvements were modest due to the limited capacity of the models used (11B). As observed in our experiments with varying dataset sizes (\textbf{\textit{n}}), the gains from augmentation plateaued beyond a certain dataset size, reflecting the model’s capacity limits. However, for larger models with greater capacity, data augmentation using memes has the potential to yield better results. This highlights the scalability and practical utility of meme-based augmentation in advancing hateful video detection, particularly in scenarios with access to high-capacity models.

\subsection{Case Studies and Error Analysis}
\label{error}

\begin{table*}
  \caption{Examples of four videos with corresponding text, ground truth (GT) labels, and predictions under five strategies. The memes dataset used is FHM, with LLaMA-3.2-11B as the model. Videos 1 and 2 are from MHC, while Videos 3 and 4 are from HateMM. Correct predictions are highlighted in blue, and incorrect ones in red.}
  
  \centering
  \small
  \begin{tabular}
  {p{2.9cm}|p{3.5cm}|p{2.8cm}|p{2.8cm}|p{3.4cm}}
  \toprule 
   & \multicolumn{2}{c}{\textbf{MHC}}  & \multicolumn{2}{c}{\textbf{HateMM}}\\ \hline
   
      \multirow{2}{*}{\textbf{Video}}
    &
    \begin{minipage}[!b]{0.1\columnwidth}
  \centering
{\includegraphics[width=3.5cm, height=3.5cm]{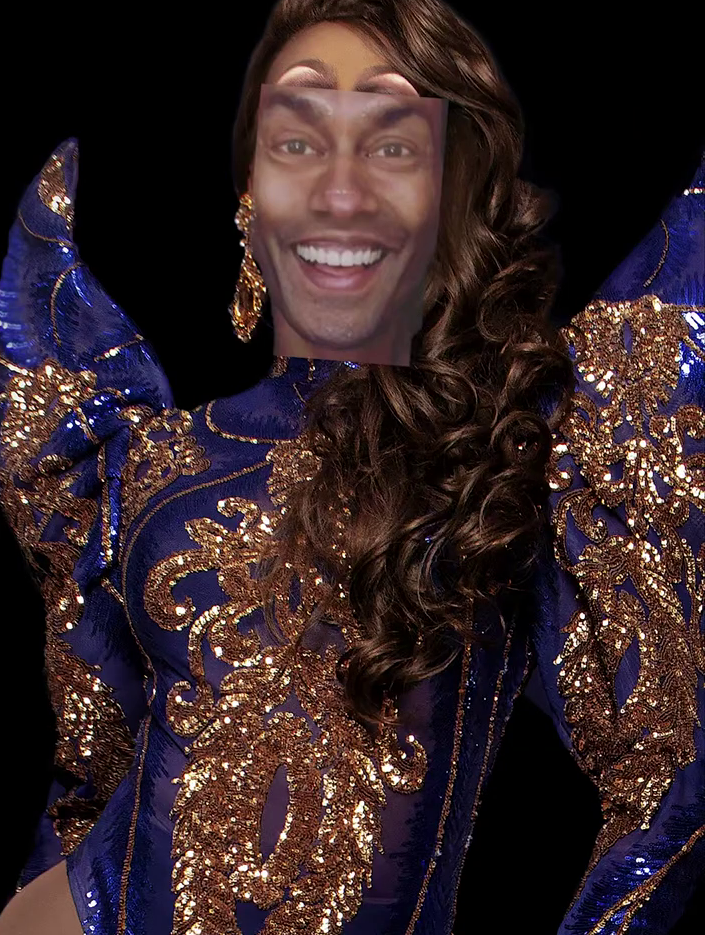}}
 \end{minipage}  
 &
    \begin{minipage}[!b]{0.1\columnwidth}
  \centering
 {\includegraphics[width=2.8cm, height=3.5cm]{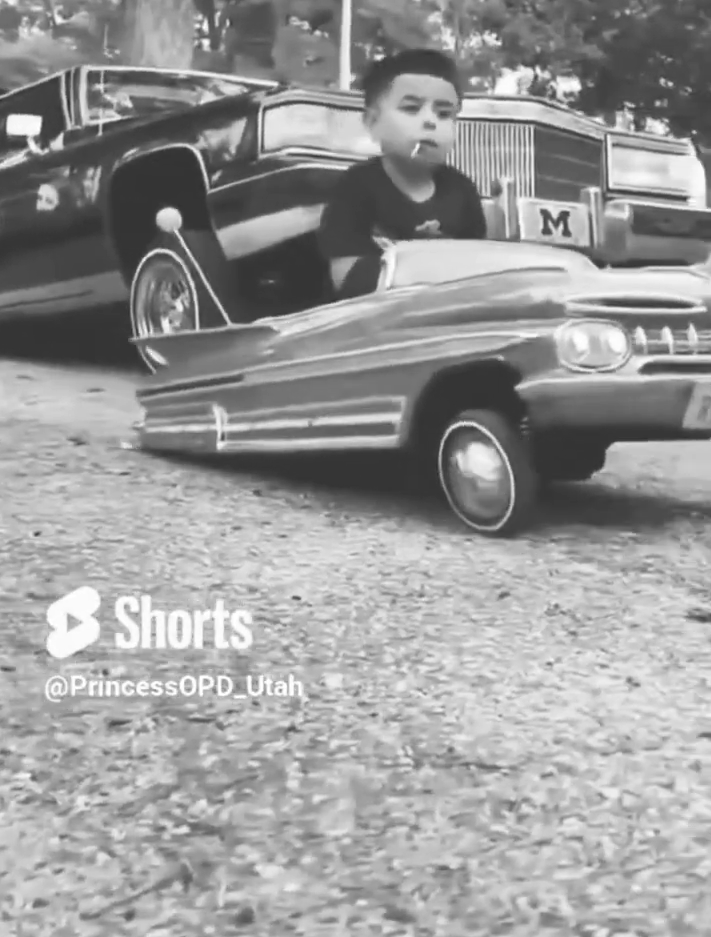}}
 \end{minipage} 
  &
    \begin{minipage}[!b]{0.1\columnwidth}
  \centering
{\includegraphics[width=2.8cm, height=3.5cm]{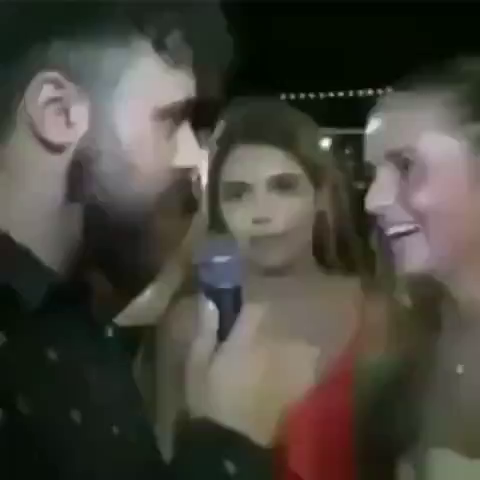}}
 \end{minipage} 
 &
    \begin{minipage}[!b]{0.1\columnwidth}
  \centering
  {\includegraphics[width=3.5cm, height=3.5cm]{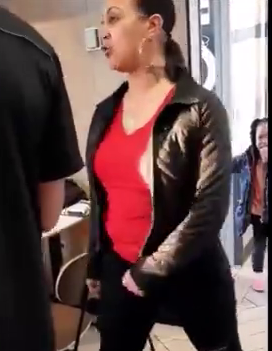}}
 \end{minipage}
 \\
   & (1) &  (2) & (3)  &  (4)\\
    \hline 
   \textbf{Text} & \textit{I m Thinking About Becoming A Man Whore. I m very poor you know I might have to prostitute myself on the streets one day...} & \textit{January you} & \textit{Is it okay to say the N word Yes No Yes No Yes No Yes Let s hear it No Let s hear it Nigga} & \textit{I ve been waiting for two weeks What the fuck is wrong with you You got my fucking daughter out here bitch Come on...} \\  \hline

     \textbf{GT} & \textbf{offensive} & \textbf{non-offensive} & \textbf{hateful} & \textbf{non-hateful} \\  \hline

     \textbf{No FT} & \textcolor{red}{non-offensive} & \textcolor{blue}{non-offensive} & \textcolor{red}{non-hateful} & \textcolor{blue}{non-hateful} \\  \hline
     
       \textbf{Vid-FT} & \textcolor{red}{non-offensive} & \textcolor{red}{offensive} & \textcolor{red}{non-hateful} & \textcolor{red}{hateful} \\  \hline
 \textbf{OM-FT} & \textcolor{red}{non-offensive} & \textcolor{blue}{non-offensive} & \textcolor{blue}{hateful} & \textcolor{blue}{non-hateful} \\  \hline
          \textbf{RM-FT} & \textcolor{blue}{offensive} & \textcolor{blue}{non-offensive} & \textcolor{blue}{hateful} & \textcolor{blue}{non-hateful} \\  \hline
          
         \textbf{Vid+RM-FT} & \textcolor{blue}{offensive} & \textcolor{blue}{non-offensive} & \textcolor{blue}{hateful} & \textcolor{blue}{non-hateful} \\  \hline

  \end{tabular}

   \label{tab:case_study}
\end{table*}

Table~\ref{tab:case_study} provides examples of four videos, detailing their corresponding text, ground truth (GT) labels, and model predictions across the five experimental strategies. The four examples are representative, ranging from video 2, which is entirely non-offensive, to video 1, which is mildly annoying and tagged as offensive, video 4, which appears violent yet lacks hateful intent, and video 3, which seems joyful but conveys hate towards Black individuals. These case studies utilize the FHM dataset for training, with LLaMA-3.2-11B as the model. The results offer valuable insights into the strengths and weaknesses of each fine-tuning strategy.

In the \textbf{No FT} setting, the model consistently predicted non-offensive or non-hateful labels, resulting in a 50\% error rate. This underscores the model’s inability to detect offensiveness or hatefulness effectively without task-specific fine-tuning. Conversely, the \textbf{Vid-FT} strategy, despite being trained directly on video datasets, misclassified all four cases, reflecting the limited generalization of video-only training. 

The \textbf{OM-FT} strategy, which uses the original meme dataset, demonstrated moderate improvement, misclassifying only Video 1. However, this error was corrected under the \textbf{RM-FT} strategy, which employs the re-annotated meme dataset. These findings highlight the importance of aligning the meme dataset’s labels with the video task’s definitions to improve cross-modal transfer performance.

The \textbf{Vid+RM-FT} strategy achieved the best performance, accurately classifying all four cases. This demonstrates the efficacy of augmenting video datasets with the re-annotated meme dataset, which provides complementary information that enhances the model’s ability to capture nuanced offensive or hateful content. These results affirm the value of re-annotated memes as a reliable augmentation resource for improving hateful video detection.

Overall, the case studies reveal the significant impact of fine-tuning strategies on model performance, emphasizing the necessity of re-annotation to align dataset labels and the advantages of leveraging memes to augment video datasets. These findings reinforce the practical utility of our proposed approach in addressing the challenges of multimodal hate speech detection.

\section{Conclusion}
This paper addresses the challenge of hateful video detection caused by the scarcity of annotated video datasets. To overcome this, we proposed leveraging hateful meme datasets as substitutes or augmentations for video datasets. Due to inconsistencies in label definitions, we developed a novel re-annotation method to align meme dataset labels with video task requirements.

Experiments showed that re-annotated meme datasets effectively substitute for video datasets, enabling comparable performance in training hateful video detection models. Augmenting video datasets with re-annotated memes further enhanced performance, achieving state-of-the-art results in hateful video classification. This approach demonstrates a viable solution to resource constraints in multimodal hate speech detection.

However, we observed that model performance plateaus with increasing dataset size, suggesting greater gains may be achievable with larger, more advanced models. Overall, our work demonstrates a scalable, cost-effective strategy for enhancing multimodal classification tasks under resource constraints.




\section*{Acknowledgement}
This research / project is supported by A*STAR under its Online Trust and Safety Research Programme (Award Grant No. S24T2TS007). Any opinions, findings and conclusions or recommendations expressed in this material are those of the author(s) and do not reflect the views of the A*STAR.

\bibliographystyle{ACM-Reference-Format}
\balance
\bibliography{ref}


\begin{thebibliography}{45}


\ifx \showCODEN    \undefined \def \showCODEN     #1{\unskip}     \fi
\ifx \showDOI      \undefined \def \showDOI       #1{#1}\fi
\ifx \showISBNx    \undefined \def \showISBNx     #1{\unskip}     \fi
\ifx \showISBNxiii \undefined \def \showISBNxiii  #1{\unskip}     \fi
\ifx \showISSN     \undefined \def \showISSN      #1{\unskip}     \fi
\ifx \showLCCN     \undefined \def \showLCCN      #1{\unskip}     \fi
\ifx \shownote     \undefined \def \shownote      #1{#1}          \fi
\ifx \showarticletitle \undefined \def \showarticletitle #1{#1}   \fi
\ifx \showURL      \undefined \def \showURL       {\relax}        \fi
\providecommand\bibfield[2]{#2}
\providecommand\bibinfo[2]{#2}
\providecommand\natexlab[1]{#1}
\providecommand\showeprint[2][]{arXiv:#2}

\bibitem[Alc{\^a}ntara et~al\mbox{.}(2020)]%
        {alcantara2020offensive}
\bibfield{author}{\bibinfo{person}{C. Alc{\^a}ntara}, \bibinfo{person}{V. Moreira}, {and} \bibinfo{person}{D. Feijo}.} \bibinfo{year}{2020}\natexlab{}.
\newblock \showarticletitle{Offensive video detection: dataset and baseline results}. In \bibinfo{booktitle}{\emph{Proceedings of the Twelfth Language Resources and Evaluation Conference}}. \bibinfo{pages}{4309--4319}.
\newblock


\bibitem[Awal et~al\mbox{.}(2021)]%
        {awal2021angrybert}
\bibfield{author}{\bibinfo{person}{Md~Rabiul Awal}, \bibinfo{person}{Rui Cao}, \bibinfo{person}{Roy Ka-Wei Lee}, {and} \bibinfo{person}{Sandra Mitrovi{\'c}}.} \bibinfo{year}{2021}\natexlab{}.
\newblock \showarticletitle{Angrybert: Joint learning target and emotion for hate speech detection}. In \bibinfo{booktitle}{\emph{Pacific-Asia conference on knowledge discovery and data mining}}. Springer, \bibinfo{pages}{701--713}.
\newblock


\bibitem[Awal et~al\mbox{.}(2023)]%
        {awal2023model}
\bibfield{author}{\bibinfo{person}{Md~Rabiul Awal}, \bibinfo{person}{Roy Ka-Wei Lee}, \bibinfo{person}{Eshaan Tanwar}, \bibinfo{person}{Tanmay Garg}, {and} \bibinfo{person}{Tanmoy Chakraborty}.} \bibinfo{year}{2023}\natexlab{}.
\newblock \showarticletitle{Model-agnostic meta-learning for multilingual hate speech detection}.
\newblock \bibinfo{journal}{\emph{IEEE Transactions on Computational Social Systems}} \bibinfo{volume}{11}, \bibinfo{number}{1} (\bibinfo{year}{2023}), \bibinfo{pages}{1086--1095}.
\newblock


\bibitem[Bertasius et~al\mbox{.}(2021)]%
        {bertasius2021space_time_attention}
\bibfield{author}{\bibinfo{person}{Gedas Bertasius}, \bibinfo{person}{Heng Wang}, {and} \bibinfo{person}{Lorenzo Torresani}.} \bibinfo{year}{2021}\natexlab{}.
\newblock \showarticletitle{Is space-time attention all you need for video understanding?}. In \bibinfo{booktitle}{\emph{Proceedings of the International Conference on Machine Learning (ICML)}}, Vol.~\bibinfo{volume}{2}. \bibinfo{pages}{4}.
\newblock


\bibitem[Cao et~al\mbox{.}(2023a)]%
        {cao2023procap}
\bibfield{author}{\bibinfo{person}{R. Cao}, \bibinfo{person}{M.~S. Hee}, \bibinfo{person}{A. Kuek}, \bibinfo{person}{W.~H. Chong}, \bibinfo{person}{R.~K.~W. Lee}, {and} \bibinfo{person}{J. Jiang}.} \bibinfo{year}{2023}\natexlab{a}.
\newblock \showarticletitle{Pro-cap: Leveraging a frozen vision-language model for hateful meme detection}. In \bibinfo{booktitle}{\emph{Proceedings of the 31st ACM International Conference on Multimedia}}. \bibinfo{pages}{5244--5252}.
\newblock


\bibitem[Cao and Lee(2020)]%
        {cao2020hategan}
\bibfield{author}{\bibinfo{person}{Rui Cao} {and} \bibinfo{person}{Roy Ka-Wei Lee}.} \bibinfo{year}{2020}\natexlab{}.
\newblock \showarticletitle{Hategan: Adversarial generative-based data augmentation for hate speech detection}. In \bibinfo{booktitle}{\emph{Proceedings of the 28th International Conference on Computational Linguistics}}. \bibinfo{pages}{6327--6338}.
\newblock


\bibitem[Cao et~al\mbox{.}(2023b)]%
        {cao2023prompting}
\bibfield{author}{\bibinfo{person}{R. Cao}, \bibinfo{person}{R.~K.~W. Lee}, \bibinfo{person}{W.~H. Chong}, {and} \bibinfo{person}{J. Jiang}.} \bibinfo{year}{2023}\natexlab{b}.
\newblock \showarticletitle{Prompting for Multimodal Hateful Meme Classification}.
\newblock \bibinfo{journal}{\emph{arXiv preprint arXiv:2302.04156}} (\bibinfo{year}{2023}).
\newblock


\bibitem[Cao et~al\mbox{.}(2020)]%
        {cao2020deephate}
\bibfield{author}{\bibinfo{person}{Rui Cao}, \bibinfo{person}{Roy Ka-Wei Lee}, {and} \bibinfo{person}{Tuan-Anh Hoang}.} \bibinfo{year}{2020}\natexlab{}.
\newblock \showarticletitle{DeepHate: Hate speech detection via multi-faceted text representations}. In \bibinfo{booktitle}{\emph{Proceedings of the 12th ACM Conference on Web Science}}. \bibinfo{pages}{11--20}.
\newblock


\bibitem[Cao et~al\mbox{.}(2024)]%
        {cao2024modularized}
\bibfield{author}{\bibinfo{person}{R. Cao}, \bibinfo{person}{R.~K.~W. Lee}, {and} \bibinfo{person}{J. Jiang}.} \bibinfo{year}{2024}\natexlab{}.
\newblock \showarticletitle{Modularized Networks for Few-shot Hateful Meme Detection}. In \bibinfo{booktitle}{\emph{Proceedings of the ACM on Web Conference 2024}}. \bibinfo{pages}{4575--4584}.
\newblock


\bibitem[Carreira and Zisserman(2017)]%
        {carreira2017temporal}
\bibfield{author}{\bibinfo{person}{João Carreira} {and} \bibinfo{person}{Andrew Zisserman}.} \bibinfo{year}{2017}\natexlab{}.
\newblock \showarticletitle{Temporal 3D ConvNets: New Architecture and Transfer Learning for Video Classification}. In \bibinfo{booktitle}{\emph{Proceedings of the IEEE International Conference on Computer Vision (ICCV)}}. \bibinfo{pages}{6568--6577}.
\newblock


\bibitem[Chhabra and Vishwakarma(2023)]%
        {chhabra2023survey}
\bibfield{author}{\bibinfo{person}{A. Chhabra} {and} \bibinfo{person}{D.~K. Vishwakarma}.} \bibinfo{year}{2023}\natexlab{}.
\newblock \showarticletitle{A literature survey on multimodal and multilingual automatic hate speech identification}.
\newblock \bibinfo{journal}{\emph{Multimedia Systems}} \bibinfo{volume}{29}, \bibinfo{number}{3} (\bibinfo{year}{2023}), \bibinfo{pages}{1203--1230}.
\newblock


\bibitem[Das et~al\mbox{.}(2023)]%
        {das2023hatemm}
\bibfield{author}{\bibinfo{person}{M. Das}, \bibinfo{person}{R. Raj}, \bibinfo{person}{P. Saha}, \bibinfo{person}{B. Mathew}, \bibinfo{person}{M. Gupta}, {and} \bibinfo{person}{A. Mukherjee}.} \bibinfo{year}{2023}\natexlab{}.
\newblock \showarticletitle{Hatemm: A Multi-Modal Dataset for Hate Video Classification}. In \bibinfo{booktitle}{\emph{Proceedings of the International AAAI Conference on Web and Social Media}}, Vol.~\bibinfo{volume}{17}. \bibinfo{pages}{1014--1023}.
\newblock


\bibitem[Davidson et~al\mbox{.}(2017)]%
        {davidson2017automated}
\bibfield{author}{\bibinfo{person}{Thomas Davidson}, \bibinfo{person}{Dana Warmsley}, \bibinfo{person}{Michael Macy}, {and} \bibinfo{person}{Ingmar Weber}.} \bibinfo{year}{2017}\natexlab{}.
\newblock \showarticletitle{Automated Hate Speech Detection and the Problem of Offensive Language}. In \bibinfo{booktitle}{\emph{Proceedings of the International AAAI Conference on Web and Social Media}}, Vol.~\bibinfo{volume}{11}. \bibinfo{pages}{512--515}.
\newblock


\bibitem[de~Gibert et~al\mbox{.}(2018)]%
        {de-gibert-etal-2018-hate}
\bibfield{author}{\bibinfo{person}{Ona de Gibert}, \bibinfo{person}{Naiara Perez}, \bibinfo{person}{Aitor Garc{\'\i}a-Pablos}, {and} \bibinfo{person}{Montse Cuadros}.} \bibinfo{year}{2018}\natexlab{}.
\newblock \showarticletitle{Hate Speech Dataset from a White Supremacy Forum}. In \bibinfo{booktitle}{\emph{Proceedings of the 2nd Workshop on Abusive Language Online ({ALW}2)}}, \bibfield{editor}{\bibinfo{person}{Darja Fi{\v{s}}er}, \bibinfo{person}{Ruihong Huang}, \bibinfo{person}{Vinodkumar Prabhakaran}, \bibinfo{person}{Rob Voigt}, \bibinfo{person}{Zeerak Waseem}, {and} \bibinfo{person}{Jacqueline Wernimont}} (Eds.). \bibinfo{publisher}{Association for Computational Linguistics}, \bibinfo{address}{Brussels, Belgium}, \bibinfo{pages}{11--20}.
\newblock
\urldef\tempurl%
\url{https://doi.org/10.18653/v1/W18-5102}
\showDOI{\tempurl}


\bibitem[Dosovitskiy et~al\mbox{.}(2021)]%
        {dosovitskiy2021vit}
\bibfield{author}{\bibinfo{person}{Alexey Dosovitskiy}, \bibinfo{person}{Lucas Beyer}, \bibinfo{person}{Alexander Kolesnikov}, \bibinfo{person}{Dirk Weissenborn}, \bibinfo{person}{Xiaohua Zhai}, \bibinfo{person}{Thomas Unterthiner}, \bibinfo{person}{Mostafa Dehghani}, \bibinfo{person}{Matthias Minderer}, \bibinfo{person}{Georg Heigold}, \bibinfo{person}{Sylvain Gelly}, \bibinfo{person}{Jakob Uszkoreit}, {and} \bibinfo{person}{Neil Houlsby}.} \bibinfo{year}{2021}\natexlab{}.
\newblock \showarticletitle{An Image is Worth 16x16 Words: Transformers for Image Recognition at Scale}. In \bibinfo{booktitle}{\emph{International Conference on Learning Representations (ICLR)}}.
\newblock


\bibitem[Dubey et~al\mbox{.}(2024)]%
        {dubey2024llama}
\bibfield{author}{\bibinfo{person}{A. Dubey}, \bibinfo{person}{A. Jauhri}, \bibinfo{person}{A. Pandey}, \bibinfo{person}{A. Kadian}, \bibinfo{person}{A. Al-Dahle}, \bibinfo{person}{A. Letman}, {and} \bibinfo{person}{R. Ganapathy}.} \bibinfo{year}{2024}\natexlab{}.
\newblock \showarticletitle{The LLAMA 3 Herd of Models}.
\newblock \bibinfo{journal}{\emph{arXiv preprint arXiv:2407.21783}} (\bibinfo{year}{2024}).
\newblock


\bibitem[Fersini et~al\mbox{.}(2022)]%
        {fersini2022semeval}
\bibfield{author}{\bibinfo{person}{Elisabetta Fersini}, \bibinfo{person}{Francesca Gasparini}, \bibinfo{person}{Giuliano Rizzi}, \bibinfo{person}{Andrea Saibene}, \bibinfo{person}{Borja Chulvi}, \bibinfo{person}{Paolo Rosso}, {and} \bibinfo{person}{Jonas Sorensen}.} \bibinfo{year}{2022}\natexlab{}.
\newblock \showarticletitle{SemEval-2022 Task 5: Multimedia Automatic Misogyny Identification}. In \bibinfo{booktitle}{\emph{Proceedings of the 16th International Workshop on Semantic Evaluation (SemEval-2022)}}. \bibinfo{pages}{533--549}.
\newblock


\bibitem[Fersini et~al\mbox{.}(2018)]%
        {fersini2018overview}
\bibfield{author}{\bibinfo{person}{Elisabetta Fersini}, \bibinfo{person}{Debora Nozza}, {and} \bibinfo{person}{Paolo Rosso}.} \bibinfo{year}{2018}\natexlab{}.
\newblock \showarticletitle{Overview of the EVALITA 2018 task on automatic misogyny identification (AMI)}. In \bibinfo{booktitle}{\emph{CEUR Workshop Proceedings}}, Vol.~\bibinfo{volume}{2263}. \bibinfo{pages}{1--9}.
\newblock


\bibitem[Gao and Huang(2017)]%
        {gao2017detecting}
\bibfield{author}{\bibinfo{person}{Lei Gao} {and} \bibinfo{person}{Ruihong Huang}.} \bibinfo{year}{2017}\natexlab{}.
\newblock \showarticletitle{Detecting online hate speech using context-aware models}.
\newblock \bibinfo{journal}{\emph{arXiv preprint arXiv:1710.07395}} (\bibinfo{year}{2017}).
\newblock


\bibitem[Hee et~al\mbox{.}(2024a)]%
        {hee2024bridging}
\bibfield{author}{\bibinfo{person}{M.~S. Hee}, \bibinfo{person}{A. Kumaresan}, {and} \bibinfo{person}{R.~K.~W. Lee}.} \bibinfo{year}{2024}\natexlab{a}.
\newblock \showarticletitle{Bridging Modalities: Enhancing Cross-Modality Hate Speech Detection with Few-Shot In-Context Learning}.
\newblock \bibinfo{journal}{\emph{arXiv preprint arXiv:2410.05600}} (\bibinfo{year}{2024}).
\newblock


\bibitem[Hee et~al\mbox{.}(2024b)]%
        {hee2024recent}
\bibfield{author}{\bibinfo{person}{M.~S. Hee}, \bibinfo{person}{S. Sharma}, \bibinfo{person}{R. Cao}, \bibinfo{person}{P. Nandi}, \bibinfo{person}{P. Nakov}, \bibinfo{person}{T. Chakraborty}, {and} \bibinfo{person}{R. Lee}.} \bibinfo{year}{2024}\natexlab{b}.
\newblock \showarticletitle{Recent Advances in Online Hate Speech Moderation: Multimodality and the Role of Large Models}. In \bibinfo{booktitle}{\emph{Findings of the Association for Computational Linguistics: EMNLP 2024}}. \bibinfo{pages}{4407--4419}.
\newblock


\bibitem[Hu et~al\mbox{.}(2021)]%
        {hu2021lora}
\bibfield{author}{\bibinfo{person}{Edward Hu}, \bibinfo{person}{Xuezhi Peng}, \bibinfo{person}{Yi Li}, \bibinfo{person}{Xifeng Liu}, \bibinfo{person}{Jie He}, \bibinfo{person}{Ziyang Chen}, \bibinfo{person}{Ziyi Li}, \bibinfo{person}{Yiming Zhang}, \bibinfo{person}{Caiming Xiong}, {and} \bibinfo{person}{Kai-Wei Chang}.} \bibinfo{year}{2021}\natexlab{}.
\newblock \showarticletitle{LoRA: Low-Rank Adaptation of Large Language Models}. In \bibinfo{booktitle}{\emph{Proceedings of the 39th International Conference on Machine Learning (ICML)}}.
\newblock
\urldef\tempurl%
\url{https://arxiv.org/abs/2106.09685}
\showURL{%
\tempurl}


\bibitem[Kiela et~al\mbox{.}(2020)]%
        {kiela2020hateful}
\bibfield{author}{\bibinfo{person}{Douwe Kiela}, \bibinfo{person}{Hamed Firooz}, {and} \bibinfo{person}{et~al. Mohan, Ankur}.} \bibinfo{year}{2020}\natexlab{}.
\newblock \showarticletitle{The Hateful Memes Challenge: Detecting Hate Speech in Multimodal Memes}.
\newblock \bibinfo{journal}{\emph{Advances in Neural Information Processing Systems}}  \bibinfo{volume}{33} (\bibinfo{year}{2020}), \bibinfo{pages}{2611--2624}.
\newblock


\bibitem[Lee et~al\mbox{.}(2024)]%
        {lee2024improving}
\bibfield{author}{\bibinfo{person}{Dong-Ho Lee}, \bibinfo{person}{Hyundong Cho}, \bibinfo{person}{Woojeong Jin}, \bibinfo{person}{Jihyung Moon}, \bibinfo{person}{Sungjoon Park}, \bibinfo{person}{Paul R{\"o}ttger}, \bibinfo{person}{Jay Pujara}, {and} \bibinfo{person}{Roy Ka-Wei Lee}.} \bibinfo{year}{2024}\natexlab{}.
\newblock \showarticletitle{Improving covert toxicity detection by retrieving and generating references}. In \bibinfo{booktitle}{\emph{Proceedings of the 8th Workshop on Online Abuse and Harms (WOAH 2024)}}. \bibinfo{pages}{266--274}.
\newblock


\bibitem[Liu et~al\mbox{.}(2022)]%
        {liu2022video_swin}
\bibfield{author}{\bibinfo{person}{Ze Liu}, \bibinfo{person}{Jia Ning}, \bibinfo{person}{Yue Cao}, \bibinfo{person}{Yixuan Wei}, \bibinfo{person}{Zheng Zhang}, \bibinfo{person}{Stephen Lin}, {and} \bibinfo{person}{Han Hu}.} \bibinfo{year}{2022}\natexlab{}.
\newblock \showarticletitle{Video swin transformer}. In \bibinfo{booktitle}{\emph{Proceedings of the IEEE/CVF Conference on Computer Vision and Pattern Recognition (CVPR)}}. \bibinfo{pages}{3202--3211}.
\newblock


\bibitem[Maity et~al\mbox{.}(2022)]%
        {maity2022multitask}
\bibfield{author}{\bibinfo{person}{Koushik Maity}, \bibinfo{person}{Pankaj Jha}, \bibinfo{person}{Subhasis Saha}, {and} \bibinfo{person}{Pawan Bhattacharyya}.} \bibinfo{year}{2022}\natexlab{}.
\newblock \showarticletitle{A multitask framework for sentiment, emotion and sarcasm aware cyberbullying detection from multi-modal code-mixed memes}. In \bibinfo{booktitle}{\emph{Proceedings of the 45th International ACM SIGIR Conference on Research and Development in Information Retrieval}}. \bibinfo{publisher}{ACM}, \bibinfo{pages}{1739--1749}.
\newblock


\bibitem[Ng et~al\mbox{.}(2024a)]%
        {ng2024love}
\bibfield{author}{\bibinfo{person}{L.~H.~X. Ng}, \bibinfo{person}{A.~X.~W. Lim}, {and} \bibinfo{person}{R.~K.~W. Lee}.} \bibinfo{year}{2024}\natexlab{a}.
\newblock \showarticletitle{Love-Hate Dataset: A Multi-Modal Multi-Platform Dataset Depicting Emotions in the 2023 Israel-Hamas War}. In \bibinfo{booktitle}{\emph{Companion Proceedings of the ACM on Web Conference 2024}}. \bibinfo{pages}{1807--1815}.
\newblock


\bibitem[Ng et~al\mbox{.}(2024b)]%
        {ng2024sghatecheck}
\bibfield{author}{\bibinfo{person}{Ri~Chi Ng}, \bibinfo{person}{Nirmalendu Prakash}, \bibinfo{person}{Ming~Shan Hee}, \bibinfo{person}{Kenny Tsu~Wei Choo}, {and} \bibinfo{person}{Roy Ka-Wei Lee}.} \bibinfo{year}{2024}\natexlab{b}.
\newblock \showarticletitle{SGHateCheck: Functional Tests for Detecting Hate Speech in Low-Resource Languages of Singapore}.
\newblock \bibinfo{journal}{\emph{arXiv preprint arXiv:2405.01842}} (\bibinfo{year}{2024}).
\newblock


\bibitem[Pan et~al\mbox{.}(2022)]%
        {pan2022st}
\bibfield{author}{\bibinfo{person}{J. Pan}, \bibinfo{person}{Z. Lin}, \bibinfo{person}{X. Zhu}, \bibinfo{person}{J. Shao}, {and} \bibinfo{person}{H. Li}.} \bibinfo{year}{2022}\natexlab{}.
\newblock \showarticletitle{ST-Adapter: Parameter-efficient Image-to-video Transfer Learning}. In \bibinfo{booktitle}{\emph{Advances in Neural Information Processing Systems (NeurIPS)}}, Vol.~\bibinfo{volume}{35}. \bibinfo{pages}{26462--26477}.
\newblock


\bibitem[Pramanick et~al\mbox{.}(2021)]%
        {pramanick2021detecting}
\bibfield{author}{\bibinfo{person}{Souvik Pramanick}, \bibinfo{person}{Dimitar Dimitrov}, \bibinfo{person}{Ritam Mukherjee}, \bibinfo{person}{Shubham Sharma}, \bibinfo{person}{Md~Shad Akhtar}, \bibinfo{person}{Preslav Nakov}, {and} \bibinfo{person}{Tanmoy Chakraborty}.} \bibinfo{year}{2021}\natexlab{}.
\newblock \showarticletitle{Detecting Harmful Memes and Their Targets}.
\newblock \bibinfo{journal}{\emph{arXiv preprint arXiv:2110.00413}} (\bibinfo{year}{2021}).
\newblock


\bibitem[Razavi et~al\mbox{.}(2010)]%
        {razavi2010offensive}
\bibfield{author}{\bibinfo{person}{Aminul-Haq Razavi}, \bibinfo{person}{Diana Inkpen}, \bibinfo{person}{Sasha Uritsky}, {and} \bibinfo{person}{Stan Matwin}.} \bibinfo{year}{2010}\natexlab{}.
\newblock \showarticletitle{Offensive Language Detection Using Multi-level Classification}.
\newblock \bibinfo{journal}{\emph{Advances in Artificial Intelligence}}  \bibinfo{volume}{6085} (\bibinfo{year}{2010}), \bibinfo{pages}{16--27}.
\newblock


\bibitem[Schmidt and Wiegand(2017)]%
        {schmidt2017survey}
\bibfield{author}{\bibinfo{person}{Anna Schmidt} {and} \bibinfo{person}{Michael Wiegand}.} \bibinfo{year}{2017}\natexlab{}.
\newblock \showarticletitle{A survey on hate speech detection using natural language processing}. In \bibinfo{booktitle}{\emph{Proceedings of the Fifth International Workshop on Natural Language Processing for Social Media}}. \bibinfo{pages}{1--10}.
\newblock


\bibitem[Wang et~al\mbox{.}(2023)]%
        {wang2023evaluating}
\bibfield{author}{\bibinfo{person}{H. Wang}, \bibinfo{person}{M.~S. Hee}, \bibinfo{person}{M.~R. Awal}, \bibinfo{person}{K.~T.~W. Choo}, {and} \bibinfo{person}{R.~K.~W. Lee}.} \bibinfo{year}{2023}\natexlab{}.
\newblock \showarticletitle{Evaluating GPT-3 generated explanations for hateful content moderation}.
\newblock \bibinfo{journal}{\emph{arXiv preprint arXiv:2305.17680}} (\bibinfo{year}{2023}).
\newblock


\bibitem[Wang et~al\mbox{.}(2024)]%
        {wang2024multihateclip}
\bibfield{author}{\bibinfo{person}{Han Wang}, \bibinfo{person}{T.~R. Yang}, \bibinfo{person}{U. Naseem}, {and} \bibinfo{person}{Roy Ka-Wei Lee}.} \bibinfo{year}{2024}\natexlab{}.
\newblock \showarticletitle{Multihateclip: A multilingual benchmark dataset for hateful video detection on YouTube and Bilibili}. In \bibinfo{booktitle}{\emph{Proceedings of the 32nd ACM International Conference on Multimedia}}. \bibinfo{pages}{7493--7502}.
\newblock


\bibitem[Warner and Hirschberg(2012)]%
        {warner2012detecting}
\bibfield{author}{\bibinfo{person}{William Warner} {and} \bibinfo{person}{Julia Hirschberg}.} \bibinfo{year}{2012}\natexlab{}.
\newblock \showarticletitle{Detecting hate speech on the World Wide Web}. In \bibinfo{booktitle}{\emph{Proceedings of the second workshop on language in social media}}. \bibinfo{pages}{19--26}.
\newblock


\bibitem[Waseem(2016)]%
        {waseem-2016-racist}
\bibfield{author}{\bibinfo{person}{Zeerak Waseem}.} \bibinfo{year}{2016}\natexlab{}.
\newblock \showarticletitle{Are You a Racist or Am {I} Seeing Things? Annotator Influence on Hate Speech Detection on {T}witter}. In \bibinfo{booktitle}{\emph{Proceedings of the First Workshop on {NLP} and Computational Social Science}}, \bibfield{editor}{\bibinfo{person}{David Bamman}, \bibinfo{person}{A.~Seza Do{\u{g}}ru{\"o}z}, \bibinfo{person}{Jacob Eisenstein}, \bibinfo{person}{Dirk Hovy}, \bibinfo{person}{David Jurgens}, \bibinfo{person}{Brendan O{'}Connor}, \bibinfo{person}{Alice Oh}, \bibinfo{person}{Oren Tsur}, {and} \bibinfo{person}{Svitlana Volkova}} (Eds.). \bibinfo{publisher}{Association for Computational Linguistics}, \bibinfo{address}{Austin, Texas}, \bibinfo{pages}{138--142}.
\newblock
\urldef\tempurl%
\url{https://doi.org/10.18653/v1/W16-5618}
\showDOI{\tempurl}


\bibitem[Waseem and Hovy(2016a)]%
        {waseem2016hateful}
\bibfield{author}{\bibinfo{person}{Zeerak Waseem} {and} \bibinfo{person}{Dirk Hovy}.} \bibinfo{year}{2016}\natexlab{a}.
\newblock \showarticletitle{Hateful Symbols or Hateful People? Predictive Features for Hate Speech Detection on Twitter}. In \bibinfo{booktitle}{\emph{Proceedings of the NAACL Student Research Workshop}}. \bibinfo{pages}{88--93}.
\newblock


\bibitem[Waseem and Hovy(2016b)]%
        {waseem-hovy-2016-hateful}
\bibfield{author}{\bibinfo{person}{Zeerak Waseem} {and} \bibinfo{person}{Dirk Hovy}.} \bibinfo{year}{2016}\natexlab{b}.
\newblock \showarticletitle{Hateful Symbols or Hateful People? Predictive Features for Hate Speech Detection on {T}witter}. In \bibinfo{booktitle}{\emph{Proceedings of the {NAACL} Student Research Workshop}}, \bibfield{editor}{\bibinfo{person}{Jacob Andreas}, \bibinfo{person}{Eunsol Choi}, {and} \bibinfo{person}{Angeliki Lazaridou}} (Eds.). \bibinfo{publisher}{Association for Computational Linguistics}, \bibinfo{address}{San Diego, California}, \bibinfo{pages}{88--93}.
\newblock
\urldef\tempurl%
\url{https://doi.org/10.18653/v1/N16-2013}
\showDOI{\tempurl}


\bibitem[Wu and Bhandary(2020)]%
        {wu2020detection}
\bibfield{author}{\bibinfo{person}{C.~S. Wu} {and} \bibinfo{person}{U. Bhandary}.} \bibinfo{year}{2020}\natexlab{}.
\newblock \showarticletitle{Detection of hate speech in videos using machine learning}. In \bibinfo{booktitle}{\emph{2020 International Conference on Computational Science and Computational Intelligence (CSCI)}}. IEEE, \bibinfo{pages}{585--590}.
\newblock


\bibitem[Xiao et~al\mbox{.}(2024)]%
        {xiao2024toxicloakcn}
\bibfield{author}{\bibinfo{person}{Yunze Xiao}, \bibinfo{person}{Yujia Hu}, \bibinfo{person}{Kenny Tsu~Wei Choo}, {and} \bibinfo{person}{Roy Ka-wei Lee}.} \bibinfo{year}{2024}\natexlab{}.
\newblock \showarticletitle{ToxiCloakCN: Evaluating Robustness of Offensive Language Detection in Chinese with Cloaking Perturbations}.
\newblock \bibinfo{journal}{\emph{arXiv preprint arXiv:2406.12223}} (\bibinfo{year}{2024}).
\newblock


\bibitem[Yao et~al\mbox{.}(2023)]%
        {yao2023side4video}
\bibfield{author}{\bibinfo{person}{H. Yao}, \bibinfo{person}{W. Wu}, {and} \bibinfo{person}{Z. Li}.} \bibinfo{year}{2023}\natexlab{}.
\newblock \showarticletitle{Side4video: Spatial-temporal side network for memory-efficient image-to-video transfer learning}.
\newblock \bibinfo{journal}{\emph{arXiv preprint arXiv:2311.15769}} (\bibinfo{year}{2023}).
\newblock


\bibitem[Zampieri et~al\mbox{.}(2019)]%
        {zampieri2019predicting}
\bibfield{author}{\bibinfo{person}{Marcos Zampieri}, \bibinfo{person}{Shervin Malmasi}, \bibinfo{person}{Preslav Nakov}, \bibinfo{person}{Sara Rosenthal}, \bibinfo{person}{Noura Farra}, {and} \bibinfo{person}{Ritesh Kumar}.} \bibinfo{year}{2019}\natexlab{}.
\newblock \showarticletitle{Predicting the Type and Target of Offensive Posts in Social Media}. In \bibinfo{booktitle}{\emph{Proceedings of the NAACL-HLT 2019, Volume 1 (Long and Short Papers)}}. \bibinfo{pages}{1415--1420}.
\newblock


\bibitem[Zhang et~al\mbox{.}(2024)]%
        {zhang2024llava}
\bibfield{author}{\bibinfo{person}{Y. Zhang}, \bibinfo{person}{B. Li}, \bibinfo{person}{H. Liu}, \bibinfo{person}{Y. Lee}, \bibinfo{person}{L. Gui}, \bibinfo{person}{D. Fu}, {and} \bibinfo{person}{C. Li}.} \bibinfo{year}{2024}\natexlab{}.
\newblock \showarticletitle{Llava-next: A Strong Zero-shot Video Understanding Model}.
\newblock \bibinfo{journal}{\emph{arXiv preprint arXiv:2407.xxxxx}} (\bibinfo{year}{2024}).
\newblock


\bibitem[Zhang et~al\mbox{.}(2018)]%
        {zhang2018detecting}
\bibfield{author}{\bibinfo{person}{Ziyi Zhang}, \bibinfo{person}{Derek Robinson}, {and} \bibinfo{person}{John Tepper}.} \bibinfo{year}{2018}\natexlab{}.
\newblock \showarticletitle{Detecting hate speech on Twitter using a convolution-GRU based deep neural network}. In \bibinfo{booktitle}{\emph{The Semantic Web: 15th International Conference, ESWC 2018, Heraklion, Crete, Greece, June 3–7, 2018, Proceedings}}, Vol.~\bibinfo{volume}{15}. \bibinfo{publisher}{Springer International Publishing}, \bibinfo{pages}{745--760}.
\newblock


\bibitem[Zhu et~al\mbox{.}(2022)]%
        {zhu2022multimodal}
\bibfield{author}{\bibinfo{person}{J. Zhu}, \bibinfo{person}{R.~K.~W. Lee}, {and} \bibinfo{person}{W.~H. Chong}.} \bibinfo{year}{2022}\natexlab{}.
\newblock \showarticletitle{Multimodal Zero-Shot Hateful Meme Detection}. In \bibinfo{booktitle}{\emph{Proceedings of the 14th ACM Web Science Conference 2022}}. \bibinfo{pages}{382--389}.
\newblock


\end{thebibliography}

\end{document}